\documentclass[conference]{IEEEtran}
\IEEEoverridecommandlockouts
\usepackage{cite}
\usepackage{amsmath,amssymb,amsfonts}
\usepackage{algorithmic}
\usepackage{graphicx}
\usepackage{textcomp}
\usepackage{xcolor}
\def\BibTeX{{\rm B\kern-.05em{\sc i\kern-.025em b}\kern-.08em
    T\kern-.1667em\lower.7ex\hbox{E}\kern-.125emX}}
\begin{document}

\title{PPMC RL Training Algorithm:  Rough Terrain Intelligent Robots through Reinforcement Learning\\
}

\author{\IEEEauthorblockN{Tamir Blum}
\IEEEauthorblockA{\textit{Department of Aerospace Engineering} \\
\textit{Tohoku University}\\
Sendai, Japan \\
Tamir@dc.tohoku.ac.jp}
\and
\IEEEauthorblockN{Kazuya Yoshida}
\IEEEauthorblockA{\textit{Department of Aerospace Engineering} \\
\textit{Tohoku University}\\
Sendai, Japan \\
Yoshida@astro.mech.tohoku.ac.jp}
}

\maketitle

\begin{abstract}
Robots will soon learn how to make decisions and control themselves, generalizing learned behaviors to unseen scenarios. In particular, AI powered robots show promise in rough environments like the lunar surface, due to the environmental uncertainties. We address this critical generalization aspect for robot locomotion in rough terrain through a training algorithm we have created called the Path Planning and Motion Control Reinforcement Learning (PPMC RL) Training Algorithm. This algorithm is coupled with any generic reinforcement learning algorithm to teach robots how to respond to user commands and to travel to designated locations on a single neural network. In this paper, we show that the algorithm works independent of the robot structure, demonstrating that it works on a wheeled rover in addition the past results on a quadruped walking robot. Further, we take several big steps towards real world practicality by introducing a rough highly uneven terrain. Critically, we show through experiments that the robot learns to generalize to new rough terrain maps, retaining a 100\% success rate. To the best of our knowledge, this is the first paper to introduce a generic training algorithm teaching generalized PPMC in rough environments to any robot, with just the use of RL. 
\end{abstract}


\section{Introduction}
As artificial intelligence (AI) progresses, we are finally gaining the ability to test past predictions of its capabilities and find real applications in a wide array of fields. One promising area for AI is in robotics, particularly for decision making and controls in rough environments, such as outer space celestial body exploration or disaster scenarios, environments that are highly unstructured and require real time information processing. In such environments, we often don't have sufficient knowledge of the environment a priori, coupled with hazardous scenarios for humans that would better be avoided if possible. Thus, we introduce the Path Planning and Motion Control Reinforcement Learning (PPMC RL) Training Algorithm, which as the name implies, teaches path planning and motion control to robots using reinforcement learning in a simulated environment.

\begin{figure}[htbp]
\centerline{\includegraphics[width=\linewidth]{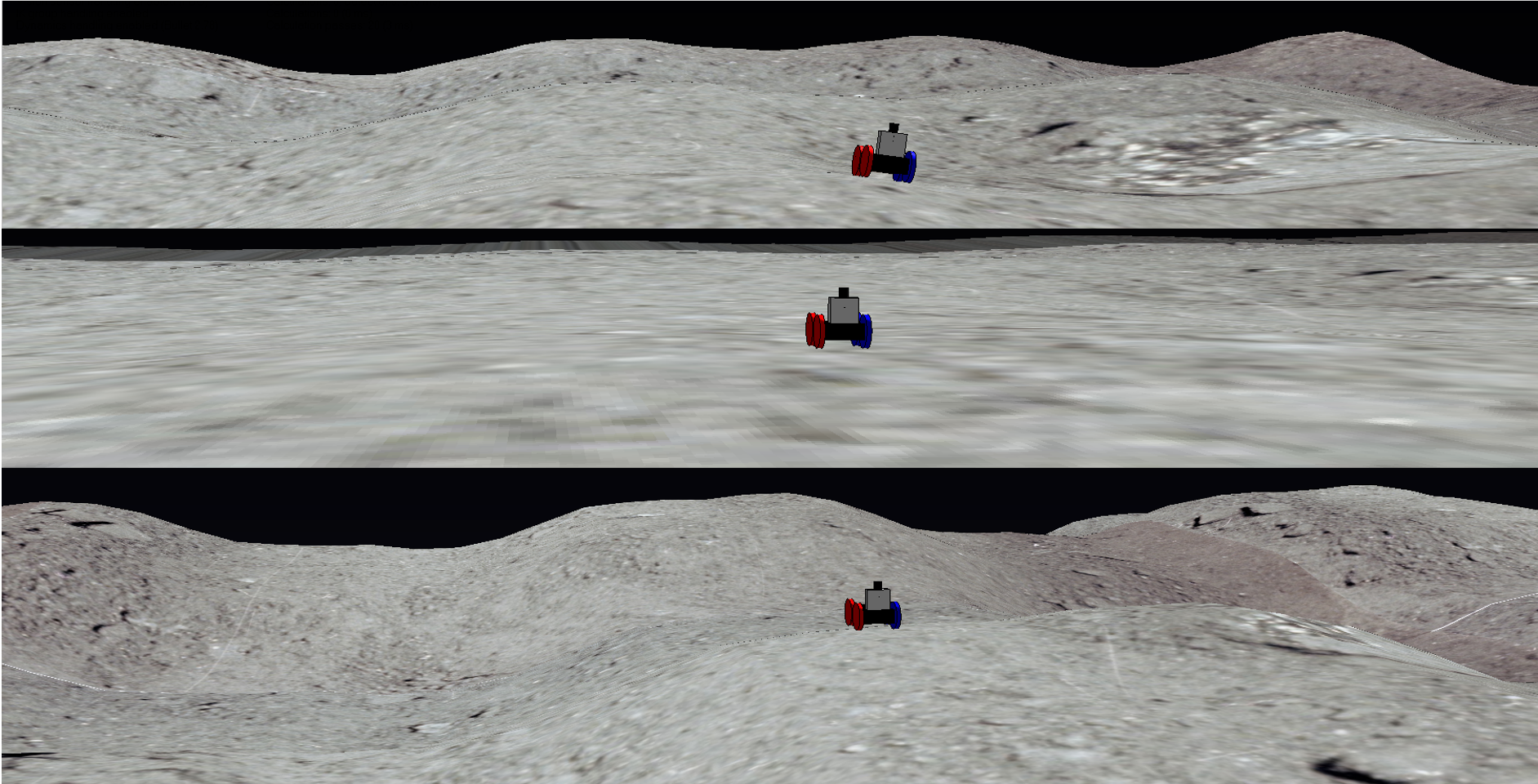}}
\caption{CLOVER in the simulated rough terrain environments emulating the lunar surface. Map 1 (the training map), Map 2 and Map 3, respectively, from top to bottom. Photo credit: NASA (lunar surface)}
\label{figmapspic}
\end{figure}

In this paper, we build upon a previously published version of the PPMC RL Training Algorithm and make several improvements that we show lead to the ability to generalize from a training map to never before seen terrains on CLOVER\cite{PPMC}\cite{clover}, a four wheeled rover platform developed by the Space Robotics Lab (SRL) for lunar surface exploration. The terrains used in this paper are now rough, bumpy terrains, contrasting with the flat terrain used in the previous work. The current training environment, Map 1, and the two other simulated environments used in our experiments, Map 2 and Map 3, as well as CLOVER, can be seen in Figure \ref{figmapspic}.

In this paper, we have taken a big step closer to simulating real world environments through the introduction of these rough and uneven surfaces. Since we don't have perfect a priori data in many real world scenarios, it is important to be able to deploy the robot in a range of environments, which could be both more and less rough than the one it was trained in. Thus, the ability to train on one environment and to generalize to different unseen environments, which could be either "harder" or "easier", is critical. We show that the PPMC RL Training Algorithm is able to achieve such generalization through experiments on the 3 maps with the CLOVER rover. 

\begin{figure}[htbp]
\centerline{\includegraphics[width=\linewidth]{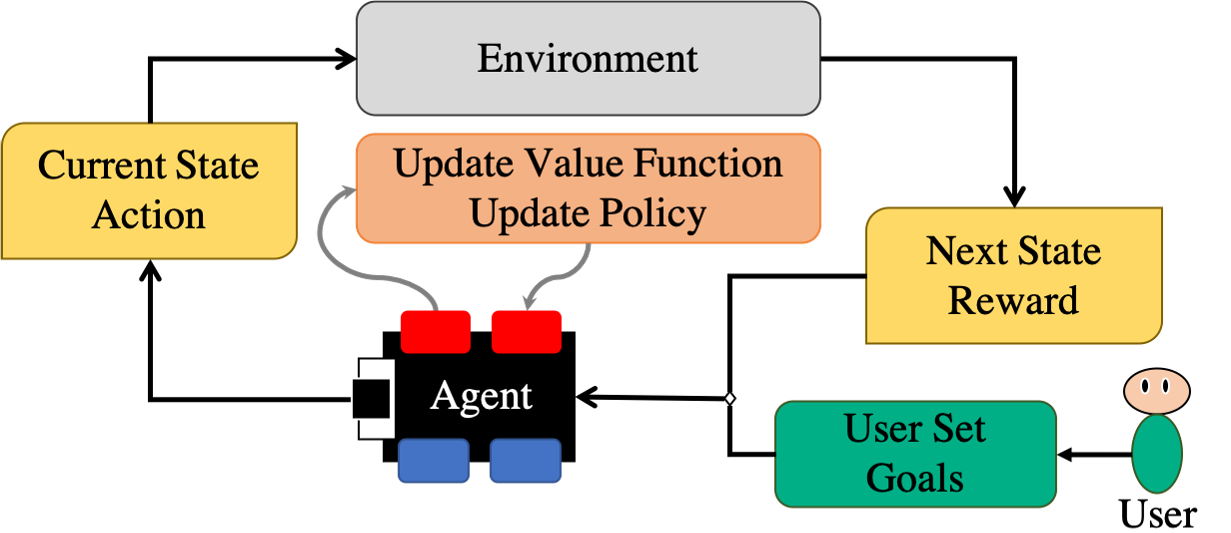}}
\caption{Ternary RL architecture: Agent – User – Environment}
\label{figUAE}
\end{figure}

This work has two main contributions, which to the best of our knowledge, are both novel. The first is the introduction of a training algorithm that can train simulated robots to conduct PPMC in rough terrain using RL alone, with the ability to generalize to new environments not trained in. Second, by demonstrating that the algorithm works on the CLOVER rover, in addition to having worked on the walking robot used in the previous work, we show that the PPMC RL Training Algorithm works independently of robot configuration or mode of locomotion. The first is important as it shows that RL has now progressed enough to be taken seriously as a viable option for robot PPMC in various real-life scenarios, such as lunar exploration or disaster response. The second is important because it can greatly reduce the amount of engineering work required to design mobile robots of any type. The RL system architecture of our setup, shown in Figure \ref{figUAE}, is notable for including a human in the loop, which was introduced in the prior PPMC publication in a novel way. 

\section{Background}
AI is a promising tool due to its ability to process large amounts of data in real time, promising to increase automation and environmental awareness. Machine learning, which is the general process of trying to learn from data, can be broken into three sub categories: supervised learning, unsupervised learning and RL. Of these, this paper focuses on RL, which we believe holds special promise for the field of robotics because of its ability to generate its own data that would otherwise be unattainable, and its ability to continuously learn by interacting with the environment. 

A RL powered robot would explore its range of possible actions and generally learns to maximize a reward function during the training. The reward function is a function comprised of variables that are specified by us before the training process. It is our way of controlling how the robot behaves, and it does this by assigning a score to each action that the robot takes, thus providing a label for the data and differentiating good actions from bad actions. In this case, we save the decision making process, called a policy, on a neural network (NN), which takes in state information about the CLOVER rover, as well as the environment and the goals. Finally, it translates these inputs into actions such as wheel speeds. The robot, alongside the policy, is called an agent.

As this paper's focus is on the improvements made to the PPMC RL Algorithm and their ramifications, we will first introduce the algorithm, along with past deficiencies and the changes we introduced. Next, we will explain the simulation conditions and the experiments before proceeding to related works, which is towards the end of the paper. Finally, we will present the explain the implications in discussion and conclusion at the end of the paper.

\section{PPMC RL Training Algorithm}
\subsection{Overview}
The PPMC RL Training Algorithm uses RL to teach a robot to conduct PPMC, and to respond to user chosen waypoints that the rover must go through. 

There are two parts to path planning and motion control. The first is path planning, which encompasses how the robot drives towards where we want it to go and recognizing that it should in fact go towards the goals. The second part motion control, which is how to control the motors for locomotion.  For this case, this encompasses controlling the two motors on the rover to create movement. The motors must be coordinated at different speeds to drive forward or backwards and turn clockwise (CW) or counterclockwise (CCW). The robot starts knowing nothing about itself, physics, the environment or its objectives. For example, the robot does not know that moving all the wheels positively causes it to go forward, that gravity pushes it down a slope, or that it has to drive towards the goal. It has to learn all this. This is known as model-free RL.

In order to be useful in real world scenarios, we want the robot to be able to learn to do PPMC in rough terrain, as shown in Figure \ref{figmapspic}. Moreover, the robot should be able to generalize what it learns to terrains of different roughness. Lastly, we would like the algorithm to reduce the overall effort required in engineering a robot capable of PPMC. Thus, we designed the training algorithm to work independently of any particular robot system architecture or mode of locomotion. 

\subsection{How the Algorithm Works}
The PPMC RL Training Algorithm works by using a combination of observable goals and randomized waypoints during the episodic training. A training perimeter is specified by the user before training, which is simply an imaginary border in the training map in which all randomized waypoints will be generated. At the start of each episode, an intermediate and a final waypoint are given to the robot and its goal is to traverse through both in the right order. Since we are using episodic training, each training episode lasts a certain duration and resets afterwards. After resetting, the robot starts again from the same origin, $(0,0)$, on the same training map. If the agent reaches the intermediate waypoint, the time limit of the current episode increases, giving the robot additional time to get to the final waypoint and fully accomplish its goal. 

As training occurs, the data is stored and used in batches to improving the policy of the robot according to the RL algorithm. Learning is done on a model-free basis, with the initial policy just containing random noise and exploring until it learns properly according to the reward function. 

\textbf{Algorithm Terminology:}
TP = training perimeter; M = number of waypoints; Alg = learning algorithm; BX, BY = boundary threshold for X and Y; tinc = episode length increase; tep = initial episode length; t = current time; e = episode counter; GX, GY = goal X and Y coordinates; PX, PY = current X and Y coordinates of robot body; R(G) = reward function w.r.t G; G = goal array; env = environment

\begin{figure}[htbp]
\label{alg:ppmc}
    \begin{algorithmic}
        \FOR{$e \in \{1,... ,N\} $}
        	\STATE Generate M randomized waypoints within TP
            \STATE Set G
        	\STATE Set R(G)
	        \STATE Run simulation with Alg, env
	        \IF{$ |P_X-G_X| \leq B_X $ \textbf{and} $ |P_Y-G_Y| \leq B_Y $}
                \STATE Update R(G)
                \STATE Update G
                \STATE tep += tinc  
            \ENDIF
            \IF{ $t \geq EL $}
                \STATE End episode, e += 1
            \ENDIF
        \ENDFOR
    \end{algorithmic}
    \caption{PPMC RL Training Algorithm}
\end{figure}

\subsection{Prior Deficiencies}
In the prototype of the PPMC RL Training Algorithm presented in the past publication, the robot trains on flat terrain, and was not tested or shown to be capable of handling any rough terrains. The training perimeter was in only one quadrant, centered around the point $(10,10)$ with the same area of $100m^2$ and in the same shape of a square. This means that the training perimeter did not include the origin and so the agent can guarantee a reward by walking towards the training perimeter. This allowed for a separation of the path planning aspect and the motion control aspect, as the robot first learns to walk blindly towards this training perimeter without needing to actually understand that it needs to go to goals. As it progresses and starts to reach the training perimeter, only then does it learn that it can further maximize reward by navigating to the designated goal and so it does so. Even given this simplification, the robot failed to achieve universally high success rate for traversing to various points within the training perimeter, let alone generalizing to the entire map or to unseen terrains. For certain areas in the training perimeter, the success rate was $0\%$ while for others it was $100\%$. Furthermore, the robot showed some limitations on its motion control capabilities. In particular, it could only turn one direction, either CCW or CW, depending on the training process. Thus, the resulting policy could perform great in the top left and center areas of the training perimeter but not on the bottom right region. Lastly, in the prior work, the algorithm was only shown to work on a single robot platform, a quadruped walking robot, thus leaving open the question of whether the algorithm can easily be used to train robots of differing system architecture and modes of locomotion. 

\subsection{Changes}
\begin{figure}[htbp]
\centerline{\includegraphics[width=\linewidth]{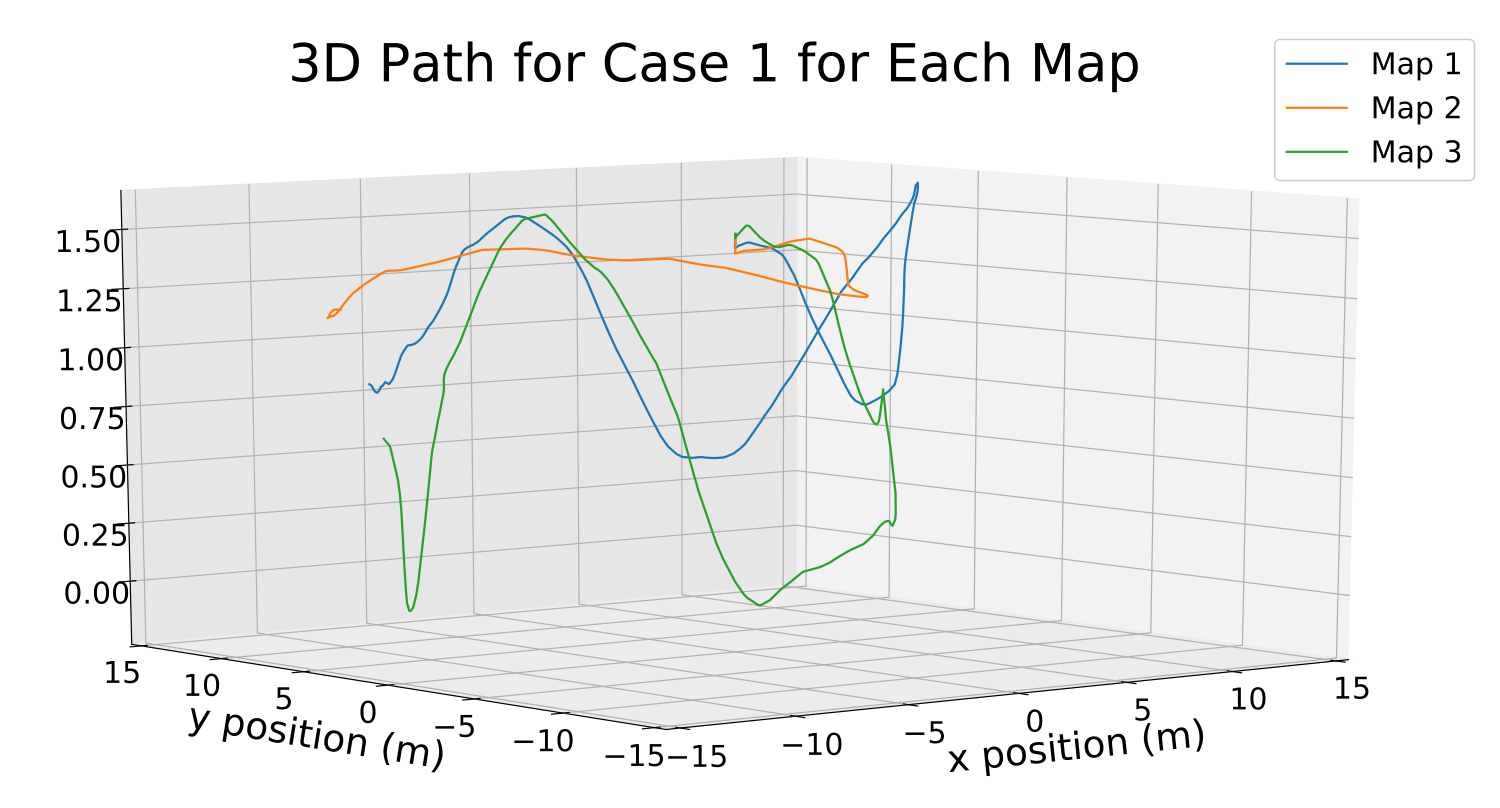}}
\caption{3D Path visualization for Case 1, in which you can notice that map 2 is smoother while map 3 is rougher than the training map, map 1.}
\label{fig3d}
\end{figure}

As for changes, first, in this paper, we create several rough terrain environments to test for the ability to traverse in rough terrain and to generalize to unseen terrains both more and less rough in nature. Their roughness can be seen visually in Figure \ref{figmapspic} or through their amplitudes shown in Figure \ref{fig3d}. In this paper we centered the training perimeter around the robot origin, while using the same square shape and same $100m^2$ total area as in the prior paper. This means we tasked the robot with waypoints in all four quadrants (in the projected Cartesian system). Further, in this work we tested the robot's ability to go both anywhere within the training region and outside. We also gave the robot access to more information and improved the data preprocessing procedure. This additional data consisted of the robot's velocity in both local and global frames, whereas in the prior paper, it was only given in the local frame. It also consisted of angular velocity, angle between robot and the current waypoint, and elapsed episode time. This resulted in a state array size to 29 elements for only two motors, whereas in the previous work it was 29 elements for 8 motors. Here, the ratio of input elements to output elements is more important than the number of elements, as a large fraction of the total state array is comprised of motor data elements ($64\%$ for the prior work, $32\%$ for the current work). In particular, we improved the data processing for angular data. Angular data, data in polar coordinates, in particular is hard for the RL agent to understand, likely due the interaction between cyclic coordinates and the way NNs interpret data. In the previous research, the angular data had two full cycles $(-2\pi,2\pi)$ within the range of $(-1,1)$.

\section{Robot-Specific Setup}
Pre-processing of data: The data should be pre-processed such that all observations are made on a scale of $(-1,1)$. This also applies for angular data. We discovered that by using linearized angular data, such as the sin and cos of the yaw angle of the robot rather than the angle itself, allows for faster training. However, it does not improve the final performance. This preprocessing must also be done for goal info. 

Post-processing of actions: The action values should be post-processed such that the range of $(-1,1)$ must be scaled to the range of useful actions for the robot. In the case of the car robot, the actions were scaled to the max motor speed going forward and in reverse.

Episode duration: The episode duration must be chosen appropriately to give the robot enough time to get to the waypoint for both near and far cases. This will heavily depend on the possible speed of the robot and the size of the training perimeter. We gave CLOVER $100s$ to get to the intermediary waypoint and doubled the time limit if CLOVER passed through the intermediate waypoint.

Fail criteria: Fail criteria should be picked such as to eliminate undesirable behaviors that the agent might take. These essentially are a harder alternative to "penalties" but could in most cases also be achieved through reward function tuning. 

State data: The state data that is passed to the agent must be chosen carefully. Within reason, the agent does not suffer by being given data it does not need, however, suffers immensely by not being given data it does need. An example is the current position of the robot, and the position of the goal, which we showed were critical to learn path planning in previous research. The form of the data, such as which reference frame is used, is also important given we use a shallow NN.

We decided to use the Actor Critic using Kronecker-Factored Trust Region (ACKTR) algorithm implemented by OpenAI’s Baselines, as it was the best of the tested algorithms in the previous work\cite{ACKTR}\cite{baselines}\cite{openaigym}.

\section{Setup}
\subsection{Reward Function}
For simplicity, we break up the reward function into 3 components, primary goal rewards, $P$, beneficial behavior rewards, $B$, and detrimental behavior penalties, $D$. The primary goal is to pass through the waypoints and so we specify this component as velocity toward the current waypoint, specified as $V_G$. For beneficial behaviors, we specify an alive reward, $A$, which encourages the robot to avoid early episode termination, triggered by certain bad behaviors. Fail conditions include excessive turning or rolling over and trigger early episode termination. Detrimental behavior is broken down into several components: a penalty for torque usage, $T$, a penalty for driving backwards, $Rv$, and a penalty for turning, $Tr$. Each term needs to be multiplied by a constant to get the right ratio of reward to penalty for each term. These constants need to be adjusted to get the proper behavior and these constants along with the terms form the reward function. Finding a good combination of weights and terms, also known as tuning the function, is probably the most time consuming part of setting up training. 

\begin{equation}
R(G) = P + B - D \label{eqrewgen}
\end{equation}
\begin{equation}
R(G) = (12.5)V_G + A/40 - Rv/100 - T/500 - Tr/18 \label{eqrew}
\end{equation}

\subsection{Simulation and Neural Network Setup}
We chose to use CoppeliaSim, a robotics simulator, due to its flexibility in designing complicated environments both for this work and for future works\cite{Coppeliasim}. We used a simulation time step of $100ms$, with bullet 2.78 selected as the physics simulator with "default" accuracy, including a $0.005s$ time step, and $100$ constraint solving iterations\cite{bullet}.

We utilized a standard feed forward NN. The training algorithm was shown in the previous work to be independent of the learning algorithm and thus there is a lot of flexibility with almost the modern day algorithms. 
 
Although the state array is the same size as the initial publication, it actually contains more information respectively, due to the reduced number of actuators in the rover as compared to the quadruped robot. This consists of
(4) motor speeds and (4) motor torques; (3) the x, y and z position of the body; (6) vectorized velocity of the body in both absolute and relative frames; (3) the angular velocity of the robot; (3) the roll, pitch and yaw orientation of the body; (1) elapsed episode time; and (1) the angle between the robot and the current waypoint. The goal array which is concatenated onto the state array remains the same size. This consists of (2) the x and y coordinate for the current goal; and (2) the x and y coordinate for the next goal. Note that motor speeds and torques are 4 elements instead of 2 elements because we included data for both the front wheel and the rear wheel even though they should always be identical.

\begin{table}[htbp]
\caption{ACKTR Learning Algorithm Parameters}
\begin{center}
\begin{tabular}{|c|c|}
\hline
\textbf{Parameter} & \textbf{\textit{Value}} \\
\hline
Batch Size & 40  \\
\hline
\# Procs & 32   \\
\hline
Learning Rate & 0.25 $\rightarrow$ 0 \\
\hline
Entropy Coeff & 0.01 \\
\hline
Clipping & 0.001  \\ 
\hline
Discounting ($\gamma$) & 0.99  \\
\hline
Value Function Coeff & 0.5 \\
\hline
\end{tabular}
\label{tabparams}
\end{center}
\end{table}

\begin{figure}[htbp]
\centerline{\includegraphics[scale=0.23]{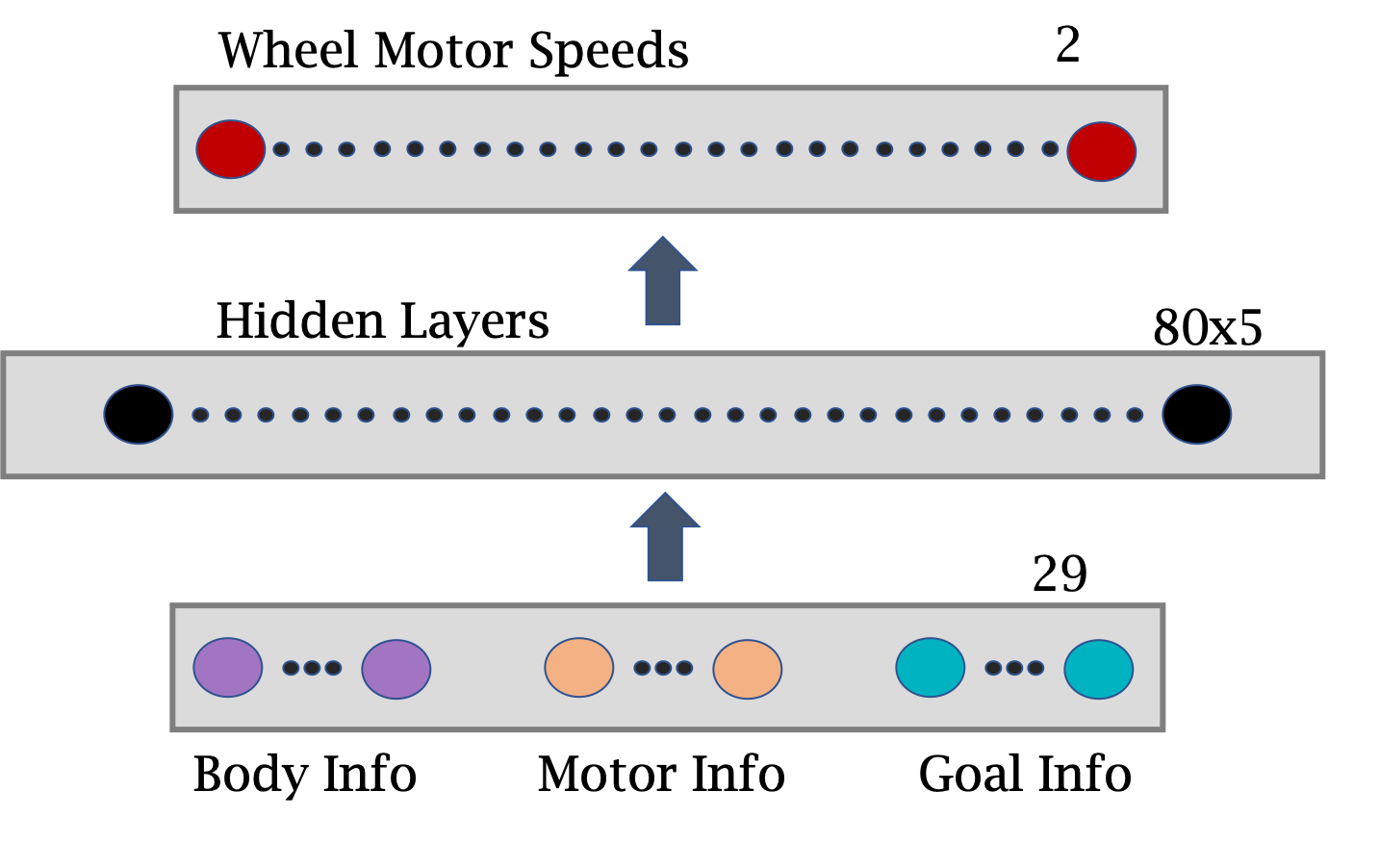}}
\caption{Simple feed forward neural network architecture used with 5 hidden layers, 80 hidden units per layer.}
\label{figNN}
\end{figure}

\subsection{Simulated Environment}
In this paper, we use a rough, uneven bumpy surface for training along with two additional maps during evaluation. These additional maps include a flatter and a bumpier variant of the training map and are meant to show that the robot agent system can generalize across different surfaces. 

\subsection{Robot}
We trained on a four wheeled rover agent in CoppeliaSim. This agent has only two motors, as its two wheel drive, yielding a 2-element action array as the output of the NN based policy, with the two left side wheels and the two right side wheels set to go the same speed, respectively. Each motor uses velocity control with a max torque of $200Nm$ and a max wheel angular velocity of $2rad/s$. The chassis is $0.3m$ long and $0.3m$ wide, and the wheels have a radius of $0.1m$. The overall robotic system weights $7.5kg$.

\section{Experiments}
We created 3 simulated environments, which we call maps, to set up some experiments to test the generalization capability of the algorithm, the general ability to conduct path planning and motion control within the rough environment, and the reliability of the robot in terms of fully accomplishing the goals we give it. 

Description of the 3 different maps, a picture of which is shown in Figure \ref{figmapspic}:
\begin{enumerate}
    \item Map 1: The training map (medium level amplitude and frequency for bumps)
    \item Map 2: Flatter variant map (decreased amplitude and frequency of bumps)
    \item Map 3: Bumpier variant map (increased amplitude and frequency of bumps)
\end{enumerate}

\begin{figure}[htbp]
\centerline{\includegraphics[width=\linewidth]{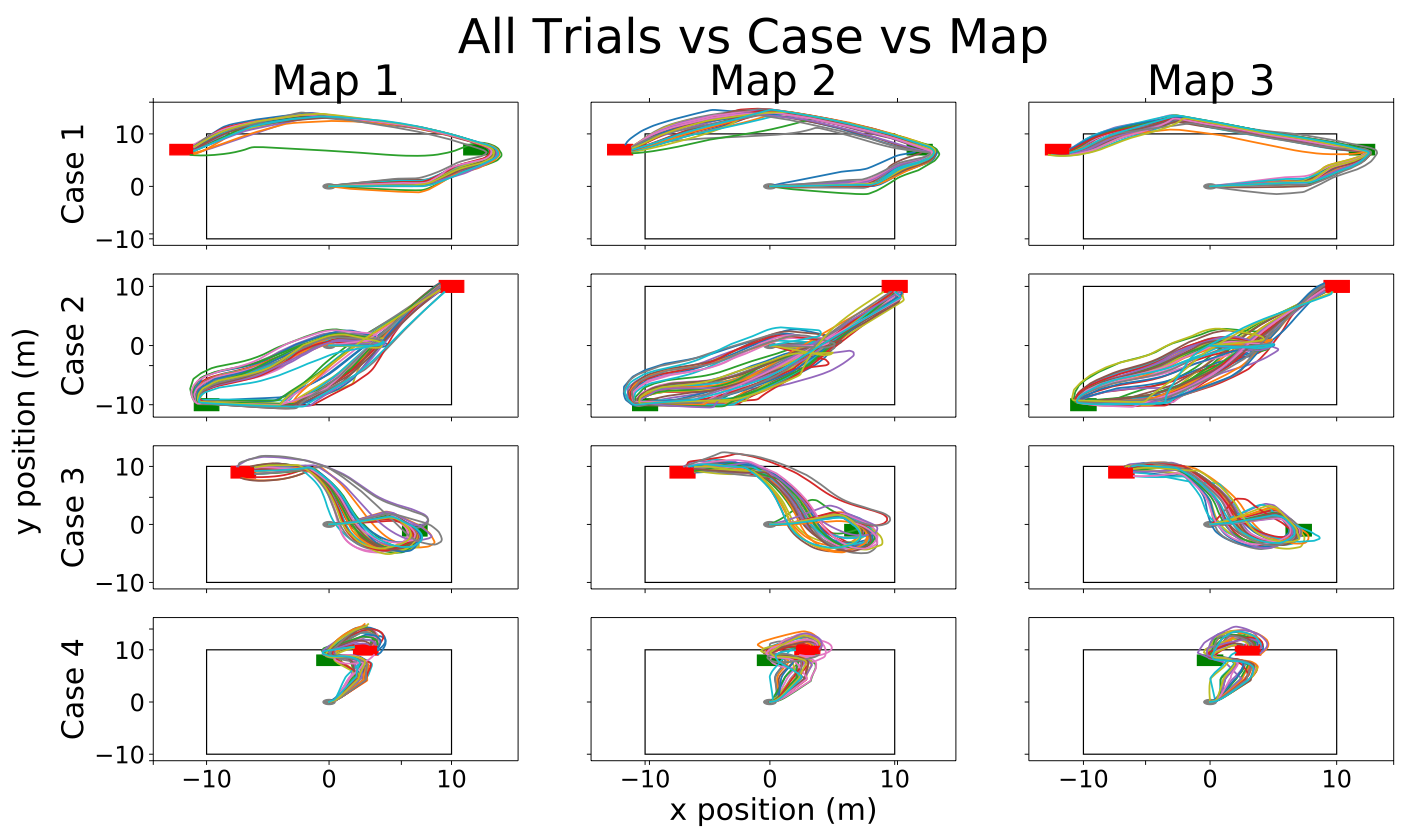}}
\caption{2D projection of all trials for all test cases for all maps are displayed. Map 1 is the training Map. The black inner rectangle is the training region. The green and red squares are the intermediate and final waypoints, respectively. Slight path deviations can be seen based on trial and based on map.}
\label{figalltrials}
\end{figure}

We created 4 test cases for the robot. Each test case consisted of an intermediate waypoint and a final waypoint with the goal being for the robot, starting at the origin, to pass through both in order. The test cases were picked to test a wide array of possibilities, including 1 test case with waypoints outside the training perimeter, 1 edge to edge test case and two other cases testing different angles of travel for both close and far waypoints. We tested the final robot-agent system on each of these 4 test cases for each of the three maps. For each test case and map combination, 30 trials were conducted, equaling a total of 12 combinations and 360 trials. 

CLOVER achieved a 100\% success rate for every test case and every map tested, as shown in Table \ref{tabsuccess}. As can be seen in figure \ref{figmapspic}, CLOVER showed fluid motion control ability, turning both CW and CCW, as appropriate.

\begin{table}[htbp]
\caption{Success ratio for two-point path test cases}
\begin{center}
\begin{tabular}{|c|c|c|c|c|}
\hline
\textbf{Test Case} & \textbf{Goal Coordinates} & \textbf{Map 1} &  \textbf{Map 2} & \textbf{Map 3} \\
\hline
1& (12,7) $\rightarrow$ (-12,7) & 100\% & 100\% & 100\%  \\
\hline
2& (-10,-10) $\rightarrow$ (10,10) & 100\% & 100\% & 100\%   \\
\hline
3 & (7,-1) $\rightarrow$ (-7,9) & 100\% & 100\% & 100\%    \\
\hline 
4 & (0,8) $\rightarrow$ (3,10) & 100\% & 100\% & 100\%    \\
\hline 
\end{tabular}
\label{tabsuccess}
\end{center}
\end{table}

\begin{figure}[htbp]
\centerline{\includegraphics[width=\linewidth]{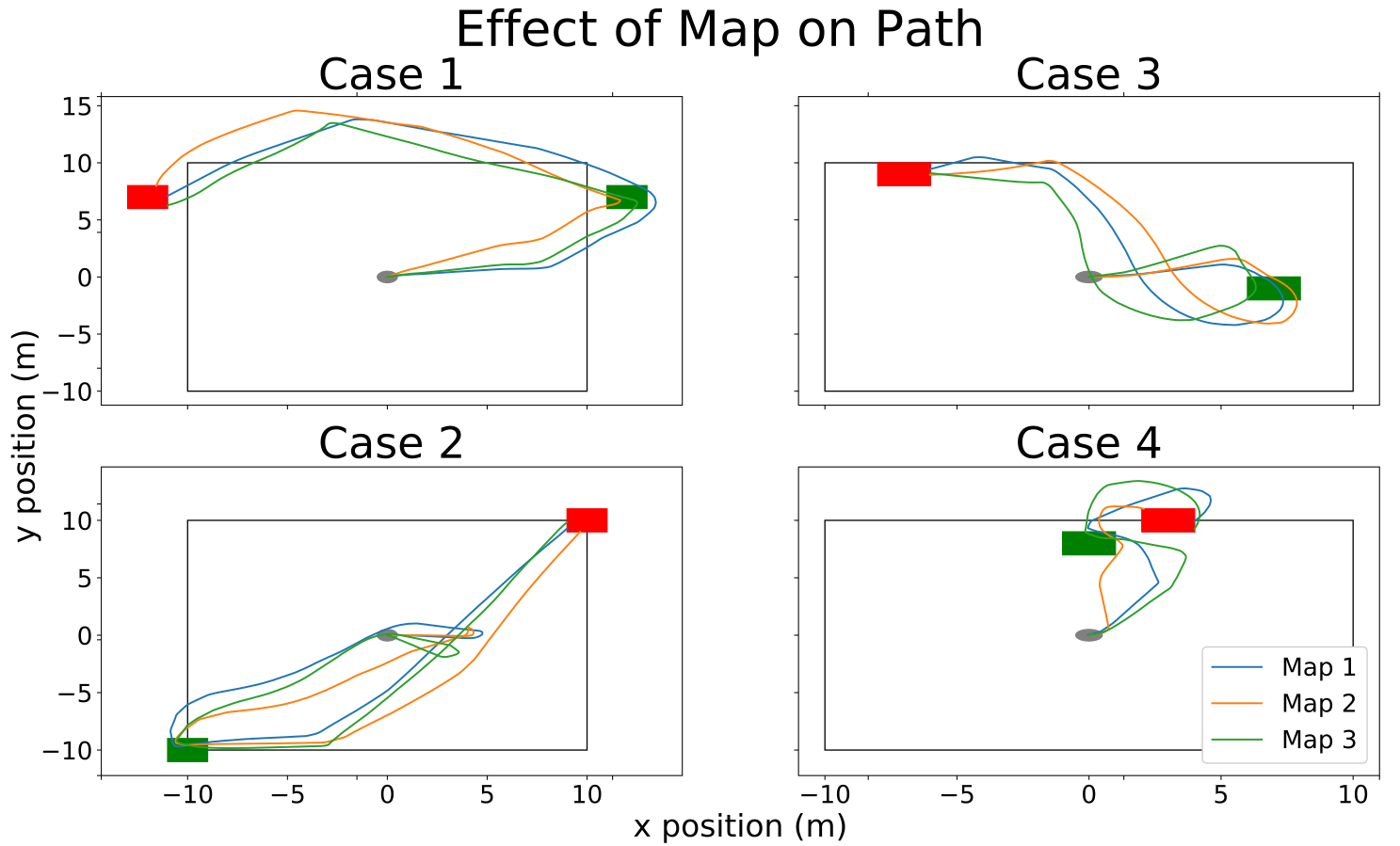}}
\caption{2D projection of CLOVER's path traversing each of the 4 test cases depending on the map.}
\label{figpath}
\end{figure}

\section{Related Work in Reinforcement Learning and Robotics}

Our work builds upon and references related work in a number of sub fields of robotics and RL. RL has been applied to robotics for some time now in areas such as motion control, mapless navigation and mapped navigation. In most cases it has been applied in simulation like in our work, however, in a limited number of other cases it has also been applied to real robots. 

There are several noteworthy works dealing with locomotion and path planning through RL. Oxford researchers were able to combine RL guided policy search with supervised learning in order to control a tensegrity rover's locomotion in bumpy terrain\cite{tensegrity}. Similar to this work, several other works have combined RL with other techniques, such as learning from expert behavior and combining a mixture of local and global path planners\cite{goaldirectednav}. HKUST researchers were able to teach a wheeled robot to path plan in a cluttered flat environment\cite{virt2realpp}. There has also been some research done focused on specializing for one particular complex environment, however without the need to generalize, allowing the agent to simply memorize actions \cite{locorichenv}\cite{williamAdaptiveSlope}. Our work uses similarly complex environments but we emphasized the need for generalization and true learning as opposed to memorization. Some of these works feature generalization in non-rough terrain, and others showcase rough terrain. We take the next step in showcasing both, and doing it through RL alone. 

There has also been research conducted by JPL researchers about current limitations and potential AI applications for mars rovers or other space exploration robots. One piece focuses on a lack of environmental awareness currently, and another about the potential energy savings that could come from the use of computer vision systems\cite{marsrover}\cite{marsvision}. This clearly shows the need for AI in space exploration rovers, and this work makes a critical step in that direction by showing the feasibility of RL based approaches. 

Some researchers have also taken a look at the human brain and the decision making process and making an analogy to complex AI systems\cite{brain}. While the referenced work is on a macro level of many agents, we are seeking to the same thing on a micro scale in our work, combining multiple functions (path planning and motion control) in a single NN on a single agent. Referencing how the brain breaks down work will become more important as we add additional systems such as vision and other sensors. 

There have been several works also combining both traditional controls with RL based controls, finding it useful for hard to model tasks with friction or for walking\cite{mixedcontrols}\cite{mixedcontrolsquad}. Our work in particular seeks to avoid this by showing it is possible using purely RL. 

\section{Discussion and Conclusion}
To our knowledge, we are the first to showcase a training algorithm that gives robots the capability to generalize in rough uneven terrain for path planning and motion control through solely RL. It is also the first RL training algorithm to be independent of robot system architecture or mode of locomotion, to the best of our knowledge. 

The improvements made in this paper are significant in terms of progressing the capabilities of RL based robots and is a step towards seeing them have uses in the real world. Having the training region and being able to generalize to the entire map and different maps with varying terrain greatly reduces the training time needed while also increasing the likelihood that the robot would work in more realistic scenarios that we cannot model perfectly, like on the lunar surface. We showed improvements to both the path planning, by achieving generalization, and the motion control, by achieving the ability to turn both CCW and CW. The complex terrains we introduced and ability to generalize will be needed if these robots are to be useful in real world applications, such as disaster aid and space exploration. High reliability will be a must in any real world system and we were able to reach a 100\% success rate for all trials conducted, even on new environments, both more and less rough. Lastly, by extending this work to a car-like rover robot, we show that it is not dependent on any particular robot system architecture. 

\begin{figure}[htbp]
\centerline{\includegraphics[width=\linewidth]{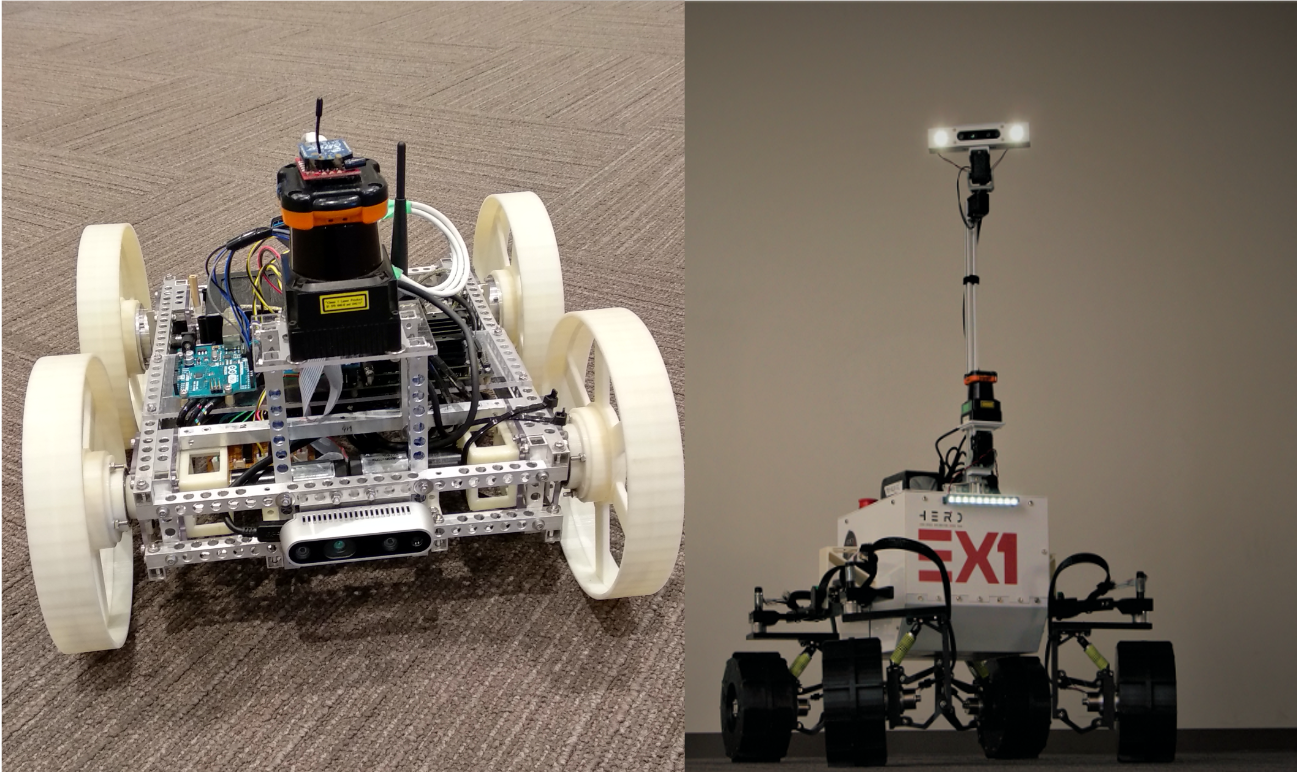}}
\caption{Two SRL Rovers, both candidates for future simulation to real world (sim2real) transfer learning work. Left, low speed swarm CLOVER prototype and right, realistic model of HERO (High-speed Exploration ROver)\cite{highspeed}.}
\label{figrovers}
\end{figure}

\bibliographystyle{IEEEtran}
\bibliography{IEEEabrv,mybib}

\end{document}